\documentclass[11pt]{article}
\usepackage[final]{acl}
\usepackage{times}
\usepackage{latexsym}
\usepackage{algorithm}
\usepackage{algorithmic}
\usepackage[T1]{fontenc}
\usepackage[utf8]{inputenc}
\usepackage{microtype}
\usepackage{inconsolata}
\usepackage{graphicx}
\usepackage{booktabs}
\usepackage{amsmath}
\usepackage{amssymb}
\usepackage{multirow}
\usepackage{subcaption}
\usepackage{url}
\usepackage{xcolor}

\title{AdaMem: Learning What to Remember with Adaptive Memory Policies for Personalized Agents}

\author{%
  Xingyu Chen \\
  Shanghai Jiao Tong University, Tencent \\
  \texttt{galaxychen@sjtu.edu.cn}
  \And
  Rui Wang$^{*}$ \\
  Shanghai Jiao Tong University \\
  \texttt{wangrui12@sjtu.edu.cn}
  \AND
  Zhaopeng Tu$^{*}$ \\
  Tencent \\
  \texttt{tuzhaopeng@gmail.com}
  \And
  Liefeng Bo \\
  Tencent \\
  \texttt{liefengbo@gmail.com}}

\begin{document}
\maketitle
\begingroup
\renewcommand{\thefootnote}{*}
\begin{NoHyper}
\footnotetext{Corresponding authors.}
\end{NoHyper}
\endgroup

\begin{abstract}
Long-term memory systems allow LLM agents to preserve information beyond a single context window, but most systems focus on storing and retrieving facts after extraction, leaving the write decision under-specified. What deserves memory can depend on the user's current task, topic, activity, or interaction partner, while uniform extraction applies one notion of importance across these different situations. We formulate this challenge as preference-conditioned write control and introduce AdaMem, which uses adaptive natural-language Memory Policies to personalize what an agent writes to memory. Each policy represents the user's memory preference for a particular interaction context, is updated from periodic feedback, and controls subsequent memory writing. We evaluate this loop in AdaMem-Bench, which assigns different memory preferences to six concurrent interaction personas across five ten-week stories. Across two extraction models and two feedback modes, AdaMem improves average QA accuracy over Mem0 from 80.0\% to 84.35\% while reducing persistent memory by 9.27\%. Our analyses show that explicit feedback helps models learn better memory policies, but current models still struggle to translate those policies into reliably selective writing behavior. AdaMem thus demonstrates the promise of adaptive write control while exposing policy execution as a central limitation of current memory agents. Our code is publicly available.\footnote{\url{https://github.com/galaxyChen/AdaMem}}
\end{abstract}

\section{Introduction}
Long-term memory is commonly used to help LLM agents maintain information across extended interactions \cite{park2023generative,wang2023survey}. Many memory systems persist conversation-derived records---including experience logs, summaries, or extracted salient information---and retrieve relevant records for later use \cite{park2023generative,packer2023memgpt,zhong2023memory,mem0}; retrieval-augmented generation provides the broader external-memory pattern \cite{lewis2020retrieval}. These architectures help agents operate beyond a single interaction context, but leave an earlier decision under-specified: \emph{which facts should be written into persistent memory in the first place?}

The decision is difficult because importance is not an intrinsic property of a fact. A user may want complete records of work activity while treating leisure-time errands as transient, or retain different information across relationships---a manager's final decisions, a partner's emotional reactions, and a colleague's technical conclusions. These examples share the same need: memory writing should follow what the user currently considers important. Uniform extraction instead fills the store with incidental details that compete with valued evidence. We call this problem \textbf{preference-conditioned write control}: personalizing what enters memory to the user's preference for a particular interaction partner or context.

\begin{figure*}[t]
  \centering
  \includegraphics[width=\textwidth]{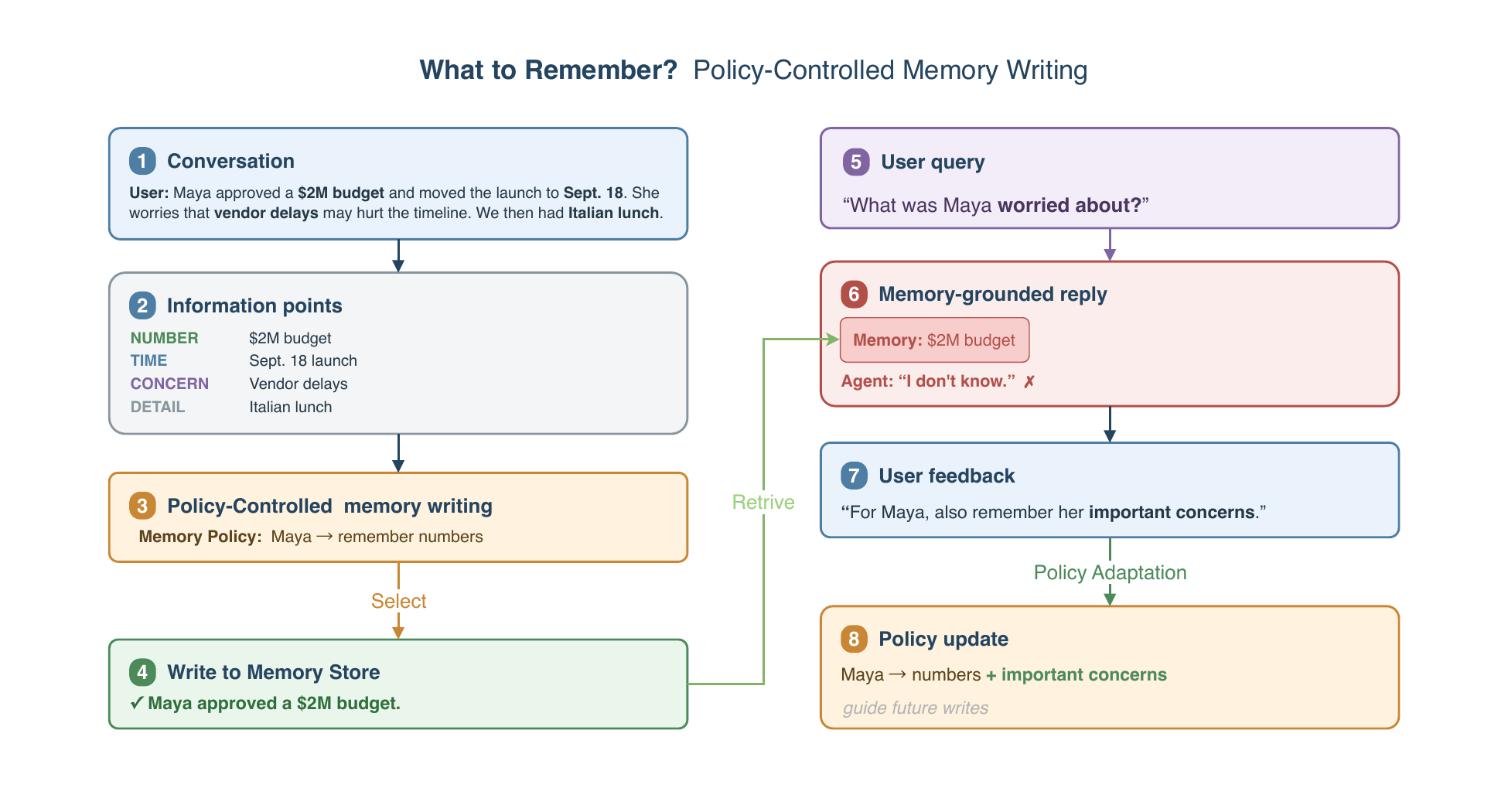}
  \caption{\textbf{Learning what to remember.} A Memory Policy controls which information is written according to the user's current preference. Feedback updates the policy, adapting future memory writing.}
  \label{fig:overview}
\end{figure*}

We introduce \textbf{AdaMem}, which equips an agent with adaptive natural-language \emph{Memory Policies} specifying what the user wants remembered in different interaction contexts. In our experiments, each policy captures the preference for a recurring interaction persona. Feedback revises these policies, which then control writing from subsequent conversations. Figure~\ref{fig:overview} illustrates this loop. To study it, we build \textbf{AdaMem-Bench}: five ten-week stories in which six recurring personas have distinct memory preferences and discuss overlapping topics, testing whether the agent retains the right information for each interaction partner.

Across different memory-extraction models and feedback modes, AdaMem improves average QA accuracy over Mem0 from 80.0\% to 84.35\% while reducing persistent memory by 9.27\%. Beyond this direct quality--compactness improvement, our analyses reveal a broader limitation of current memory agents: learning a better textual policy does not reliably produce more selective future writing. Even when explicit feedback improves the learned policy, the model rejects only about one fifth of planned distractors. We call this the \textbf{learning-to-execution gap}, a meta-control challenge in using learned rules to regulate the agent's own future memory operations.

\paragraph{Contributions.}
\begin{itemize}
  \item We formulate long-term memory construction as preference-conditioned write control and introduce AdaMem, which learns adaptive natural-language Memory Policies from feedback to personalize subsequent memory writing.
  \item We study this problem through role-specific memory preferences and build AdaMem-Bench, a longitudinal benchmark with five ten-week stories, six concurrent personas per story, overlapping topics, preference-aligned memories, and periodic feedback.
  \item Across different memory-extraction models and feedback modes, we show that AdaMem improves answer quality with a smaller persistent memory in this setting and uncover a learning-to-execution bottleneck: current models can learn better memory rules yet still fail to apply them selectively.
\end{itemize}

\section{Related Work}
\paragraph{Long-term memory construction and management.}
Retrieval-augmented and agent memory systems provide language models with external knowledge or persistent experience records \cite{lewis2020retrieval,park2023generative,packer2023memgpt,zhong2023memory,mem0,memoryos2025}. Recent methods organize memories into evolving structures \cite{amem2025}, distill or filter retained content \cite{nemori2025,memorai2026}, and learn memory-management operations \cite{agemem2026,memoryr12026}. AdaMem instead holds the storage and retrieval backend fixed and adapts a natural-language policy that controls what the extractor writes.

\paragraph{Personalization and longitudinal evaluation.}
Personalized agents adapt retrieval, responses, actions, or memory management using user-specific histories and preference representations \cite{modarressi2023retrieval,emgrag2024,personaagent2025,personamem2025,affectivemem2025,pamu2026}. Longitudinal benchmarks evaluate long-term QA and reasoning across extended, multi-session, or multi-party interactions \cite{maharana2024lococmo,wu2024longmemeval,evermembench2026}; Memora additionally targets consolidation and obsolete-memory handling \cite{memora2026}. AdaMem instead personalizes the writing interface: user feedback causes the agent to adapt its Memory Policy, which then controls what the agent retains in different interaction contexts.

\paragraph{Reflection and adaptive control.}
LLM agents use evaluative or self-generated feedback to guide later attempts \cite{shinn2023reflexion,madaan2023self}; memory methods similarly use reflective summarization and retrieval refinement \cite{prospectretro2025}, automated architecture evolution \cite{evolvemem2026}, or learned management actions \cite{agemem2026,memoryr12026}. AdaMem instead turns feedback into a persistent, human-readable rule for the earlier write decision. Separating the policy updater from the later extractor lets us evaluate both whether feedback induces the intended rule and whether the extractor follows it.

\section{AdaMem}
\subsection{Design Principles}
Our goal is to control memory writing while allowing the agent to revise its extraction behavior from observed user feedback. AdaMem realizes this goal through three principles. First, \textbf{preferences act before storage}: the system should use the user's memory preferences to select which facts enter persistent memory, rather than attempting to repair an indiscriminate store only at retrieval time. Second, \textbf{the control rule is adaptive and scope-conditioned}: feedback revises a natural-language Memory Policy---not model parameters---for a particular task, topic, activity, or interaction partner; our benchmark uses recurring personas as scopes. Third, \textbf{policy learning and policy execution are separated}: a policy updater learns or revises the rule from feedback, whereas the memory extractor executes that rule on later conversations. Evaluating these two components separately lets us determine whether a failure results from learning an incorrect policy or from failing to follow a correctly learned policy.

\subsection{Formalization}
We begin with the common workflow of long-term memory agents. The dialogue stream is divided into extraction batches $C_1,C_2,\ldots$. At extraction step $i$, an extractor $E$ converts the current batch into memories and adds them to the persistent store $\mathcal{D}$:
\begin{align}
  \mathcal{M}_i &= E(C_i), \\
  \mathcal{D} &\leftarrow \operatorname{Store}(\mathcal{D},\mathcal{M}_i).
\end{align}
When the user sends a new query $q$, the agent retrieves relevant memories and uses them to produce a response $y$:
\begin{align}
  \mathcal{R}_q &= \operatorname{Retrieve}(\mathcal{D},q), \\
  y &= A(q,\mathcal{R}_q),
\end{align}
where $A$ is the response model. This extract--store--retrieve--respond pattern covers the memory systems compared in our experiments.

AdaMem maintains a collection of scope-specific Memory Policies; below, $\mathcal{P}$ denotes the policy applicable to the current conversation. It changes the extraction step by adding $\mathcal{P}$ as an input:
\begin{equation}
  \mathcal{M}_i = E(C_i,\mathcal{P}).
\end{equation}
The Memory Policy describes the user's memory preferences, and the extractor applies those preferences when deciding which facts in $C_i$ to write. Storage, retrieval, and response generation retain the same interfaces as in the common workflow.

AdaMem also updates the Memory Policy from interaction history. Let $H_j$ be the user--assistant interaction records accumulated for the $j$-th policy update. These records can include ordinary dialogue as well as direct corrections or explanations from the user. A policy updater $U$ maps the current policy and this history to the next policy:
\begin{equation}
  \mathcal{P}_{j+1}=U(\mathcal{P}_j,H_j).
\end{equation}
The updated policy then guides subsequent extraction batches. This creates a feedback loop in which user interactions first reveal memory preferences and later influence what enters memory.

Algorithm~\ref{alg:adamem} summarizes the complete workflow. Relative to the common memory-agent pattern, AdaMem supplies the current Memory Policy to each extraction call and periodically updates that policy from accumulated interaction history.

\begin{algorithm}[t]
\caption{Adaptive Memory-Policy Learning and Execution Loop}
\label{alg:adamem}
\begin{algorithmic}[1]
\REQUIRE Dialogue stream, initial policy $\mathcal{P}$, memory store $\mathcal{D}$
\STATE initialize extraction buffer $C\leftarrow\varnothing$
\STATE initialize policy-update history $H\leftarrow\varnothing$
\FOR{each user input $q$}
  \STATE $\mathcal{R}\leftarrow\operatorname{Retrieve}(\mathcal{D},q)$
  \STATE $y\leftarrow A(q,\mathcal{R})$
  \STATE append the completed interaction $(q,y)$ to $C$ and $H$
  \IF{a policy-update boundary is reached}
    \STATE $\mathcal{P}\leftarrow U(\mathcal{P},H)$
    \STATE $H\leftarrow\varnothing$
  \ENDIF
  \IF{a memory-extraction boundary is reached}
    \STATE $\mathcal{M}\leftarrow E(C,\mathcal{P})$
    \STATE $\mathcal{D}\leftarrow\operatorname{Store}(\mathcal{D},\mathcal{M})$
    \STATE $C\leftarrow\varnothing$
  \ENDIF
  \STATE return response $y$
\ENDFOR
\end{algorithmic}
\end{algorithm}

\section{AdaMem-Bench}
\subsection{Task Format}
AdaMem-Bench evaluates preference-conditioned writing in a longitudinal memory-QA task. It contains five ten-week stories, each following six recurring interaction personas. Every week contains ten dialogue sessions followed by ten QA items, yielding 100 sessions and 100 questions per story. The agent incrementally writes the conversations to memory and answers questions about preference-aligned facts introduced earlier in the story. The benchmark additionally provides the preference and memory annotations needed to examine the write-control process.

\subsection{Data Generation and Annotations}
We construct each story with an outline-first pipeline, beginning with six recurring personas and a distinct memory preference for each, such as a manager's final decisions or an engineering colleague's technical conclusions and key numbers.

The pipeline then generates a ten-week outline organizing recurring topics, preference-aligned target facts, and related distractors, and expands each scenario into a multi-turn conversation. Alongside every conversation, it produces one or more \emph{Golden Memories}: the minimal atomic facts supported by the dialogue that match the focal persona's memory preference.

Finally, weekly QA is generated from the Golden Memories with reference answers and supporting-memory links. A stage-wise audit achieves a 93.5\% macro-average pass rate, and human review judges 93.3\% of a stratified sample of 150 decisions reasonable (Appendix~\ref{app:qc}).

\subsection{Design for Memory Write Control}
\paragraph{Competing information.}
Different personas repeatedly discuss shared or related topics while contributing both preference-aligned facts and topically related distractors. This creates semantically similar competitors during extraction and retrieval, testing whether the agent can use persona-specific preferences to select target information before it enters memory.

\paragraph{Preference identification and policy updates.}
Weekly QA interactions provide the history for policy updates. The user confirms correct answers and supplies the correct answer after an error. In the \textbf{explicit-feedback} mode, the correction also directly states the relevant memory preference. In the \textbf{implicit-feedback} mode, it omits this preference statement, so the agent must infer the recurring target information types from the questions and corrections. Thus, both modes provide answer-level correction, but only explicit feedback directly identifies what the user wants remembered.

\paragraph{Write-control supervision.}
AdaMem-Bench provides three complementary forms of supervision. Reference answers evaluate downstream QA utility. Authored persona preferences serve as references for judging whether an updated Memory Policy captures the intended rule. At the session level, Golden Memories identify preference-aligned facts that should be written, while planned distractors identify information that should be omitted; these annotations support Key Recall and Noise Rejection measurements on subsequent extraction sessions. Together, the annotations separately evaluate policy learning and policy execution.

\begin{table*}[t]
\centering
\small
\setlength{\tabcolsep}{4.2pt}
\begin{tabular}{llrrrrrrr}
\toprule
\textbf{Extractor} & \textbf{Feedback} &
\multicolumn{4}{c}{\textbf{QA Accuracy}} &
$\boldsymbol{\Delta}$ &
\multicolumn{2}{c}{\textbf{Memory Volume}} \\
\cmidrule(lr){3-6}\cmidrule(lr){8-9}
& & \textbf{Full Ctx.} & \textbf{Ideal Mem.} & \textbf{Mem0} & \textbf{AdaMem} &
& \textbf{Mem0} & \textbf{AdaMem} \\
\midrule
DeepSeek-V4-Flash & Explicit & 96.4 & 96.0 & 76.2 & 85.2 & +9.0 & 429.8 & 371.4 \\
DeepSeek-V4-Flash & Implicit & 97.2 & 96.4 & 80.2 & 82.0 & +1.8 & 407.4 & 380.0 \\
Gemini-3.5-Flash & Explicit & 96.4 & 96.0 & 80.4 & 84.8 & +4.4 & 378.4 & 351.6 \\
Gemini-3.5-Flash & Implicit & 97.2 & 96.4 & 83.2 & 85.4 & +2.2 & 391.0 & 354.6 \\
\midrule
\multicolumn{2}{l}{Average} & 96.8 & 96.2 & 80.0 & 84.35 & +4.35 & 401.7 & 364.4 \\
\bottomrule
\end{tabular}
\caption{Main results on AdaMem-Bench. Full Context and Ideal Memory are extractor-independent reference settings and are repeated across extractors for readability. Accuracy differences are AdaMem minus Mem0 in percentage points; volumes are mean final numbers of persistent memory items across five stories.}
\label{tab:matrix5}
\end{table*}

\section{Experimental Setup}
\subsection{Methods}
We use \textbf{Mem0} as the primary baseline because it implements the standard pipeline of extraction, persistent storage, retrieval, and memory-grounded answering \cite{mem0}. \textbf{AdaMem} retains this pipeline and adds an adaptive Memory Policy at extraction, with a separate rule for each learned persona. We also include two reference settings: \textbf{Full Context} supplies the complete accumulated dialogue at answer time, while \textbf{Ideal Memory} stores only pipeline-generated Golden Memories. This design isolates preference-conditioned write control while holding the remaining memory workflow fixed.

\subsection{Main Evaluation}
All main experiments use AdaMem-Bench, yielding 500 QA items in each memory-extractor and feedback-mode condition. We test DeepSeek-V4-Flash and Gemini-3.5-Flash as the memory-extraction and policy-update model, each under explicit and implicit feedback. DeepSeek-V4-Flash generates all QA answers; following the LLM-as-a-Judge paradigm \cite{zheng2023judging}, a separate method-blind DeepSeek-V4-Flash call judges answer correctness.

\subsection{Extraction and Update Cadence}\label{sec:cadence}
Memory extraction and policy updating use separate cadences. Each benchmark week contains ten dialogue sessions distributed across several dated observation days; dialogue memory is extracted once per observation date, jointly processing its assigned sessions. At the end of each week, the system answers ten QA items and receives feedback, after which AdaMem updates the Memory Policy once from the ten QA--feedback interactions. The complete QA--answer--feedback transcript is then written in one additional extraction call using the updated policy; Mem0 performs the same writeback without policy updating. Both methods therefore use identical dialogue batching and writeback boundaries.

\subsection{Metrics}
The primary metric is \textbf{QA Accuracy}, which measures final answer correctness. We report \textbf{Memory Volume}, the number of persistent memory items, to quantify memory compactness. To connect the final QA result to the main memory pipeline, we additionally report \textbf{Extraction F1} against the preference-aligned Golden Memories and \textbf{Recall@5} for whether target evidence appears among the top five retrieved memories.

\section{Results}
\subsection{Main Results: Overall Effectiveness}
We compare AdaMem with Mem0 across two memory-extraction models and two feedback modes, using Full Context and Ideal Memory as the reference points defined above.

Table~\ref{tab:matrix5} shows the complete comparison. Averaged over the four conditions, AdaMem improves QA accuracy from 80.0\% to 84.35\% (+4.35 points) and reduces the mean persistent memory from 401.7 to 364.4 items (9.27\%). The gain is largest for the DeepSeek extractor with explicit feedback and remains positive in all four conditions. Full Context reaches 96.4\%/97.2\% accuracy under explicit/implicit feedback, while Ideal Memory reaches 96.0\%/96.4\%, indicating substantial remaining room between current extraction and the two reference settings.

Table~\ref{tab:quality5} traces the result through the memory pipeline. AdaMem improves both Extraction F1 and Recall@5 in all four conditions. The F1 gains range from 1.9 to 6.4 points, indicating that the smaller persistent store is not produced by indiscriminately discarding information: the written memories match the preference-aligned targets more closely. Recall@5 rises by 4.0--10.4 points, showing that this improvement survives retrieval and makes supporting evidence more accessible to the answer model. The largest joint improvement again occurs with DeepSeek and explicit feedback, where Extraction F1 increases by 5.1 points and Recall@5 by 10.4 points. 
\begin{table*}[t]
\centering
\small
\begin{tabular}{llrrrr}
\toprule
\textbf{Extractor} & \textbf{Feedback} &
\multicolumn{2}{c}{\textbf{Extraction F1}} &
\multicolumn{2}{c}{\textbf{Recall@5}} \\
\cmidrule(lr){3-4}\cmidrule(lr){5-6}
& & \textbf{Mem0} & \textbf{AdaMem} & \textbf{Mem0} & \textbf{AdaMem} \\
\midrule
DeepSeek-V4-Flash & Explicit & 35.0 & 40.1 & 54.2 & 64.6 \\
DeepSeek-V4-Flash & Implicit & 33.9 & 40.3 & 58.2 & 68.0 \\
Gemini-3.5-Flash & Explicit & 42.4 & 47.2 & 64.2 & 73.4 \\
Gemini-3.5-Flash & Implicit & 45.1 & 47.0 & 67.8 & 71.8 \\
\bottomrule
\end{tabular}
\caption{Extraction and retrieval quality on AdaMem-Bench. F1 and Recall@5 are percentages.}
\label{tab:quality5}
\end{table*}

\subsection{Memory Write-Control Behavior}\label{sec:writecontrol}
The main results show that adding a Memory Policy changes writing and improves downstream QA, but they do not reveal which part of the adaptive loop succeeds or fails. We therefore separate two questions central to preference-conditioned write control: whether feedback produces a policy that represents the intended preference, and whether the extractor follows that policy on later conversations. To keep this behavioral analysis independent of the models used for extraction and answer judging, we use a third model family, GLM-5.2, to audit the AdaMem-Bench DeepSeek trajectories. \textbf{Policy Alignment} measures agreement between a learned persona rule and the authored preference, while \textbf{Substantive Update Direction} asks whether an operational policy change moves toward or away from that preference. \textbf{Key Recall} and \textbf{Noise Rejection} then measure whether later extraction retains planned preference-aligned facts and omits planned distractors.

\begin{figure*}[t]
  \centering
  \includegraphics[width=0.92\textwidth]{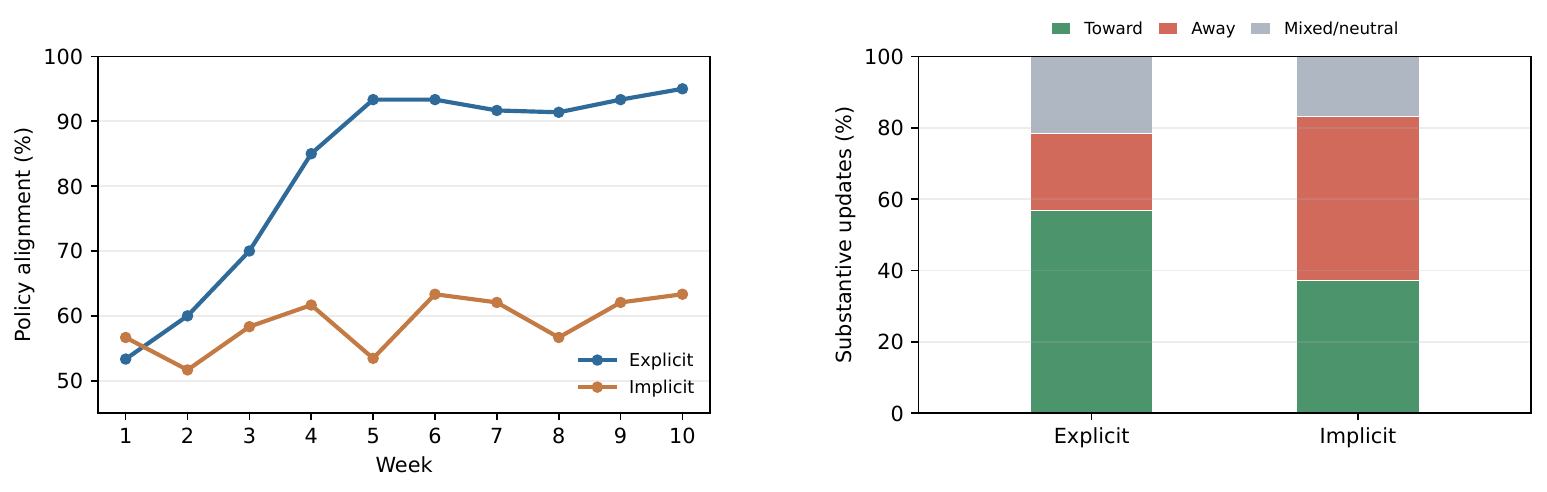}
  \caption{Policy dynamics on AdaMem-Bench with the DeepSeek extractor. Left: semantic alignment with the authored persona preference. Right: direction of substantive policy updates across scored snapshots.}
  \label{fig:policy}
\end{figure*}

\paragraph{Policy learning.}
Figure~\ref{fig:policy} shows that explicit feedback raises normalized alignment from 53.3\% in Week 1 to 95.0\% in Week 10. Implicit feedback reaches 63.3\% and continues to modify the policy, but its update direction is less reliable. Among substantive updates, 56.8\% of explicit-feedback changes move toward the intended preference, compared with 37.1\% under implicit feedback; 46.1\% of substantive implicit updates move away from it. Direct preference statements therefore produce more reliable policy learning than inferring preferences from question patterns alone.

This contrast clarifies what the two feedback modes test. Explicit feedback supplies the desired abstraction directly, so the updater can consolidate repeated corrections into a stable persona rule. Implicit feedback contains evidence only through the distribution of questions and corrections; the model must infer the latent preference and is more likely to introduce an overly broad or misdirected rule. The continued policy changes under implicit feedback therefore indicate adaptation, but not necessarily adaptation in the intended direction.

\begin{table}[t]
\centering
\small
\resizebox{\columnwidth}{!}{%
\begin{tabular}{llrr}
\toprule
\textbf{System} & \textbf{Feedback} & \textbf{Key Recall} & \textbf{Noise Rejection} \\
\midrule
Mem0 & Explicit & 88.9 & 10.9 \\
AdaMem & Explicit & 93.6 & 20.1 \\
Mem0 & Implicit & 86.6 & 10.2 \\
AdaMem & Implicit & 92.3 & 18.1 \\
\bottomrule
\end{tabular}
}
\caption{Memory-writing behavior under a common session-level audit (\%). Key Recall measures retention of planned preference-aligned facts; Noise Rejection measures omission of planned distractors.}
\label{tab:policy}
\end{table}

\paragraph{Policy execution.}
To test whether the learned policy changes later writing, we audit the memories produced by Mem0 and AdaMem on the same subsequent conversation sessions. Using the session-level annotations, we measure the fraction of Golden Memory facts retained (\textbf{Key Recall}) and the fraction of planned distractors omitted (\textbf{Noise Rejection}). As shown in Table~\ref{tab:policy}, under explicit feedback AdaMem raises Key Recall from 88.9\% to 93.6\% (+4.7 points) and Noise Rejection from 10.9\% to 20.1\% (+9.2 points). Under implicit feedback, the corresponding gains are +5.7 points in Key Recall (86.6\% to 92.3\%) and +7.9 points in Noise Rejection (10.2\% to 18.1\%). Preference-conditioned writing therefore preserves more target facts while omitting more planned distractors, confirming that the smaller AdaMem store is not explained by indiscriminate deletion.

The absolute values expose an important remaining limitation. AdaMem retains over 92\% of planned key facts, but still rejects only 18--20\% of distractors. Thus, the policy has a much stronger effect on preserving desired information than on suppressing topically related noise. Coupled with the strong improvement in Policy Alignment, this residual asymmetry constitutes the \textbf{learning-to-execution gap}: the model learns a substantially better rule and executes it better than Mem0's unguided extraction, but still does not apply the rule selectively enough. Effective write control therefore requires both accurate policy induction and stronger instruction following at extraction time.

\paragraph{Retrieval consequences.}
We further annotate the frozen top-10 retrieval results in the DeepSeek/explicit condition by whether each retrieved item contains the target evidence and which interaction persona it concerns. Appendix Table~\ref{tab:retrieval} reports the full attribution results. AdaMem raises target-evidence Hit@5 from 54.2\% to 64.6\%, showing that improved writing makes the evidence required for answering more accessible. However, 35.2\% of its top-five retrieved items concern a persona other than the one asked about; these are \emph{wrong-person memories}, often retrieved because different personas discuss overlapping topics. Thus, preserving preferred facts improves evidence recall, while incomplete suppression of competing information leaves substantial headroom relative to Ideal Memory.

\subsection{Robustness of the Main Comparison}
\paragraph{Cross-family judging.}
We freeze all 4,000 Mem0 and AdaMem answers and rejudge them with Gemini-3.5-Flash using the same method-blind correctness rubric. The two QA Judges agree on 98.45\% of answers ($\kappa=0.946$), and the average AdaMem--Mem0 difference changes only from +4.35 to +4.15 points. Thus, AdaMem's QA improvement remains statistically distinguishable from zero under a Judge from a different model family, supporting the robustness of the main comparison.

\paragraph{Scale and mixed-use robustness.}
To test generalization beyond the main five stories, we scale the benchmark to 20 ten-week stories in the DeepSeek-V4-Flash/explicit condition. AdaMem retains a positive average QA difference under both the original DeepSeek Judge (+1.45 points) and independent Gemini rejudging (+1.55 points); Appendix Table~\ref{tab:twenty} reports the complete results. We also test robustness in a more realistic mixed-use setting, where a personal assistant receives routine requests unrelated to later personal-memory questions. We simulate this setting by interleaving fixed instruction--response pairs at 1$\times$--5$\times$ the original dialogue volume while keeping the personal conversations and insertion schedule identical across methods. Across all traffic levels, AdaMem stores 13.7--18.8\% fewer memories and improves Recall@5 by 11.0--13.0 points, while its QA difference ranges from $-0.2$ to +7.2 points. Full results appear in Appendix Table~\ref{tab:dolly}. These evaluations support the compactness and retrieval benefits across larger-scale and mixed-use settings.

\section{Conclusion}
We formulate memory construction as \emph{preference-conditioned write control} and introduce AdaMem, which adapts natural-language Memory Policies from feedback to personalize future writing. To evaluate this process, AdaMem-Bench combines longitudinal QA with annotations of persona-specific preferences, target memories, and distractors. Across this benchmark, AdaMem improves average QA accuracy over Mem0 from 80.0\% to 84.35\% while reducing persistent memory by 9.27\%. Moreover, its smaller store preserves and retrieves target evidence more effectively. Nevertheless, models learn better policies while still struggling to reject preference-irrelevant information, revealing a learning-to-execution gap. Taken together, our results establish adaptive write control as a distinct component of personalized memory agents and selective policy execution as a key direction for future work.

\clearpage
\section{Limitations}
AdaMem-Bench is a synthetic, bounded evaluation with six recurring preference families. Its experiments organize preferences by interaction persona, while real users exhibit greater linguistic and behavioral diversity. Memory preferences can also be organized around tasks, topics, activities, or combinations of these conditions. Extending the Memory Policy to these settings requires detecting which preferences apply, combining simultaneously applicable rules, resolving conflicts and hierarchical priorities, and retiring preferences when a task or context ends. The limited number of story trajectories constrains statistical generalization even though every story contains ten weeks, six concurrent personas, and 100 QA items. QA scoring, mechanism labels, and most data-quality checks rely on LLM Judges; cross-family scoring and human calibration provide additional checks while leaving residual Judge dependence. Finally, the current policy update accepts valid structured patches directly. Reliable online validation remains difficult when user feedback is sparse, preferences evolve, and historical QA is an imperfect proxy for current utility.

\section*{Acknowledgements}
This work was conducted while Xingyu Chen was an intern at Tencent and was supported by the Tencent Rhino-Bird Research Program (Project No.~Tencent JR2026TEG042).

\bibliography{custom}

\appendix
\section{Scaling to More Stories}\label{app:twenty}
\paragraph{Research question.}
The main matrix uses five long stories to compare multiple extractors and feedback modes. We additionally ask whether the positive AdaMem--Mem0 direction persists across a broader set of complete trajectories in one representative configuration.

\paragraph{Experimental design.}
We evaluate the DeepSeek-extractor/explicit-feedback condition on a separately generated, predetermined set of 20 ten-week stories. Each story contains 100 QA items and six recurring interaction personas. Both methods are evaluated on all 20 stories without selecting trajectories by their observed outcome. We report the original method-blind DeepSeek Judge and an independent Gemini rejudgment of the same 4,000 frozen answers. Confidence intervals are obtained by bootstrapping stories.

\begin{table}[t]
\centering
\small
\resizebox{\columnwidth}{!}{%
\begin{tabular}{lrrrrr}
\toprule
\textbf{QA Judge} & \textbf{Mem0} & \textbf{AdaMem} & $\boldsymbol{\Delta}$ & \textbf{95\% CI} & \textbf{Wins} \\
\midrule
DeepSeek & 81.75 & 83.20 & +1.45 & [-0.35, 3.25] & 12/20 \\
Gemini & 83.35 & 84.90 & +1.55 & [0.25, 2.85] & 11/20 \\
\bottomrule
\end{tabular}}
\caption{Twenty-story replication in the DeepSeek/explicit condition. Accuracy and intervals are percentages and percentage points; Wins counts stories where AdaMem has higher QA accuracy.}
\label{tab:twenty}
\end{table}

\paragraph{Results and conclusion.}
As shown in Table~\ref{tab:twenty}, AdaMem retains a positive average difference under both Judges: +1.45 points with DeepSeek and +1.55 points with Gemini. The effect is nevertheless heterogeneous across stories, with AdaMem ahead on 12/20 and 11/20 stories, respectively; the DeepSeek-Judge confidence interval also includes zero. This replication supports the direction of the main comparison in the representative setting, while showing that its magnitude is smaller and less uniform than the five-story DeepSeek/explicit result. We therefore interpret the main matrix as evidence for an average benefit in the controlled benchmark rather than a guarantee of improvement on every trajectory.

\section{Mixed-Instruction-Traffic Stress Test}\label{app:dolly}
\paragraph{Research question.}
Personal memory systems may receive routine assistant requests that are unrelated to the later personal-memory questions. We test whether preference-conditioned writing remains useful when such traffic is interleaved with the longitudinal conversations.

\paragraph{Experimental design.}
We form a separate five-story stress test by inserting instruction--response pairs sampled from Databricks Dolly-15k \cite{conover2023dolly} between personal-chat sessions. The 1$\times$, 3$\times$, and 5$\times$ conditions add approximately one, three, and five times the original dialogue-token volume. Within each traffic condition, the personal conversations, Golden Memories, QA items, sampled instructions, and insertion schedule are fixed across methods. This isolates how Mem0 and AdaMem handle the same mixed-use stream.

\begin{table*}[t]
\centering
\small
\begin{tabular}{crrrrrrr}
\toprule
\textbf{Traffic} & \textbf{Ideal} & \textbf{Full Ctx.} & \textbf{Mem0} & \textbf{AdaMem} & $\boldsymbol{\Delta}$ & \textbf{Mem. red.} & \textbf{R@5 gain} \\
\midrule
1$\times$ & 96.0 & 95.6 & 82.2 & 87.6 & +5.4 & 14.2\% & +13.0 \\
3$\times$ & 96.0 & 95.6 & 85.8 & 85.6 & -0.2 & 13.7\% & +12.4 \\
5$\times$ & 94.0 & 96.0 & 81.0 & 88.2 & +7.2 & 18.8\% & +11.0 \\
\bottomrule
\end{tabular}
\caption{Mixed-instruction-traffic results. Accuracy and Recall@5 differences are percentage points; memory reduction is relative to Mem0.}
\label{tab:dolly}
\end{table*}

\paragraph{Results and conclusion.}
Table~\ref{tab:dolly} shows that Ideal Memory and Full Context remain at 94.0--96.0\% accuracy, indicating that the inserted traffic does not remove the evidence needed for the original QA. AdaMem stores 13.7--18.8\% fewer memories than Mem0 and improves Recall@5 by 11.0--13.0 points at every traffic level. Its QA difference is +5.4, $-0.2$, and +7.2 points at 1$\times$, 3$\times$, and 5$\times$, respectively. The robust effects are therefore a smaller persistent store and better target-evidence retrieval; the final QA gain varies with the sampled traffic composition. This stress test supports preference-conditioned writing under mixed use without implying uniform answer-quality gains.

\section{Data Quality Control}\label{app:qc}
The data audit follows the generation sequence and applies an independent rubric at each stage. Table~\ref{tab:qc} reports story-macro pass rates. The overall rate is 93.5\%. A stratified human calibration sample includes both automatically passed and flagged units; 140 of 150 automated decisions are judged reasonable.

\begin{table}[t]
\centering
\small
\begin{tabular}{lr}
\toprule
\textbf{QC target} & \textbf{Pass rate} \\
\midrule
Preference semantics/person assignment & 100.0\% \\
Outline targets, distractors, timeline & 95.8\% \\
Dialogue realization and naturalness & 95.2\% \\
Golden Memory support and coverage & 83.1\% \\
Question and reference-answer evidence & 99.3\% \\
Corrective-feedback consistency & 97.4\% \\
\midrule
Overall macro average & 93.5\% \\
Human calibration reasonableness & 93.3\% \\
\bottomrule
\end{tabular}
\caption{Stage-wise data-quality audit on AdaMem-Bench.}
\label{tab:qc}
\end{table}

The lower Golden Memory pass rate in Table~\ref{tab:qc} is driven mainly by set-level coverage and redundancy judgments, which are stricter than checking whether an individual target fact is supported. The quality-control process evaluates generated data and does not inspect Mem0 or AdaMem predictions.

\section{Retrieval Attribution}\label{app:retrieval}
We annotate the frozen top-10 retrieval results in the DeepSeek-V4-Flash/explicit-feedback condition by whether each retrieved memory contains the target evidence and whether it concerns the interaction persona named in the question. Table~\ref{tab:retrieval} reports the resulting attribution measures.

\begin{table}[t]
\centering
\small
\begin{tabular}{lrr}
\toprule
\textbf{Metric} & \textbf{Mem0} & \textbf{AdaMem} \\
\midrule
Target-Persona P@1 & 53.4\% & 60.0\% \\
Target-Persona P@5 & 36.4\% & 38.2\% \\
Wrong-Person@5 & 31.1\% & 35.2\% \\
Same-Anchor Wrong-Person@5 & 30.2\% & 33.7\% \\
Target Evidence Hit@5 & 54.2\% & 64.6\% \\
\bottomrule
\end{tabular}
\caption{Persona and target-evidence attribution of retrieved memories. Wrong-Person@5 is the proportion of the top-five retrieved items that concern a persona other than the one named in the question; Same-Anchor Wrong-Person@5 restricts these items to the same topic anchor as the query.}
\label{tab:retrieval}
\end{table}

\end{document}